\documentclass{isprs} % isprs class modified 23-04-2019 (Dennis Wittich)

\usepackage{setspace}
\usepackage{geometry} % added 27-02-2014 Markus Englich
\usepackage[labelsep=period]{caption}  % added 14-04-2016 Markus Englich - Recommendation by Sebastian Brocks
\usepackage[british]{babel} 
\usepackage[hang]{footmisc}
\usepackage[authoryear]{natbib}

\usepackage[utf8]{inputenc} % allow utf-8 input
\usepackage[T1]{fontenc}    % use 8-bit T1 fonts
\usepackage{hyperref}       % hyperlinks
\usepackage{url}            % simple URL typesetting
\usepackage{amssymb}
\usepackage{booktabs}       % professional-quality tables
\usepackage{amsfonts}       % blackboard math symbols
\usepackage{nicefrac}       % compact symbols for 1/2, etc.
\usepackage{microtype}      % microtypography
\usepackage{xcolor}         % colors
\usepackage{xspace}
\usepackage{enumitem}
\usepackage{graphicx}
\usepackage{multirow}
\usepackage{pifont}%
\usepackage{bm}
\usepackage{amsthm}
\usepackage{amsmath}
\usepackage[capitalize]{cleveref}

\usepackage{mathtools}

\newcommand{\bb}[1]{\mathbf{#1}}

\newcommand{\bx}{\bb{x}}
\newcommand{\bxi}{\bx^{(i)}}

\newcommand{\bz}{\bb{z}}

\newcommand{\bT}{\boldsymbol{\theta}}

\newcommand{\bphi}{\boldsymbol{\phi}}

\newcommand{\pT}{p_{\bT}}

\newcommand{\qPhi}{q_{\bphi}}

\newcommand{\Exp}[2]{\mathbb{E}_{#1}\left[#2\right]}

\newcommand{\eqnr}{\addtocounter{equation}{1}\tag{\theequation}}

\theoremstyle{definition}

\newcommand{\LB}[2]{\mathcal{L}^{#1}(\bT,\bphi; #2)}

\usepackage{subcaption}  % For subfigures

\newcommand{\method}{BetterScene\xspace}

\begin{document}

\title{\method: 3D Scene Synthesis with Representation-Aligned Generative Model}
\date{}

% KAO: Remove extra spacing

% Anonymous submissions, authors' names should not be visible
\author{Yuci Han, Charles Toth, John, E. Anderson, William J. Shuart, Alper Yilmaz}
% \author{***** (for review, names must be rendered anonymous)}

% KAO: Remove extra newline
% Anonymous submissions, authors' affiliations should not be visible
%\address{
%	\textsuperscript{1 }ITU, Civil Engineering Faculty, 80626 Maslak Istanbul, Turkey - (oaltan, tozg, kulur, seker)@itu.edu.tr\\
%	\textsuperscript{2 }Dept.\ of Geomatic Engineering, University College London, Gower Street, London, WC1E 6BT UK - idowman@ge.ucl.ac.uk\\
%	\textsuperscript{3 }Université Côte d’Azur, INRIA – Sophia-Antipolis, France – florent.lafarge@inria.fr\\
%	\textsuperscript{4 }Univ. Gustave Eiffel, IGN-ENSG, LaSTIG – Saint-Mandé, France – clement.mallet@ign.fr\\
%	\textsuperscript{5 }Institute of Photogrammetry and GeoInformation, Leibniz Universit\"at Hannover, Germany - heipke@ipi.uni-hannover.de\\
%}
% \address{**** (for review, affiliations must be rendered anonymous)}
\address{Dept. of Electrical and Computer Engineering, The Ohio State University\\(han.1489, toth.2, yilmaz.15)@osu.edu \\ USACE ERDC GRL, \\\{john.e.anderson, William.j.shuart\}@usace.army.mil}

% If the corresponding author is NOT the final author, always add a % space before the subsequent comma, i.e.
% first author name\textsuperscript{a,}\thanks{Corresponding author} , % second author name \textsuperscript{b}, etc.
% thanks to Niclas Borlin 05-05-2016
% information on the corresponding author should not be used any longer and has been commented out
% C. Heipke, Jan 03,2024

% the use of the information of commissions and working groups should not be used any longer and has been commented out
% C. Heipke, Sept. 20,2022
%\commission{XX, }{YY} %This field is optional. If filled, XX and YY should be replaced by adequate numbers. See https://www2.isprs.org/commissions/
%\workinggroup{XX/YY} %This field is optional.
%\icwg{}   %This field is optional.

% \author{
% Yuci Han\textsuperscript{1,2} \quad Alper Yilmaz\textsuperscript{1,2} \vspace{3pt} \\
% \textsuperscript{1}The Ohio State University \vspace{3pt} \\
% \quad
% \textsuperscript{2}Photogrammetry Computer Vision Lab \vspace{3pt} 
% \quad \\
%   \texttt{\small \{han.1489, yilmaz.15\}@osu.edu}
% }

% KAO: Use times symbol
\abstract{

We present \method, an approach to enhance novel view synthesis (NVS) quality for diverse real-world scenes, using extremely sparse unconstrained photos. \method leverages the production-ready Stable Video Diffusion (SVD) model pretrained on billions of frames as a strong backbone, aiming to mitigate artifacts and recovering view-consistent details at inference time. Conventional methods have developed similar diffusion-based solutions to address these challenges of novel view synthesis. Despite significant improvements, these methods typically rely on off-the-shelf pretrained diffusion priors and fine-tune only the UNet module while keeping other components frozen, which still leads to inconsistent details and artifacts even when incorporating geometry-aware regularizations like depth or semantic conditions. To address this, we investigate the latent space of the diffusion model and introduce two components: (1) temporal equivariance regularization and (2) vision foundation model-aligned representation, both applied to the variational autoencoder (VAE) module within the SVD pipeline. \method integrates a feed-forward 3D Gaussian Splatting (3DGS) model to render features as inputs for the SVD enhancer and generate continuous, artifacts-free, consistent novel views. We perform evaluation using the challenging DL3DV-10K dataset, demonstrating significant visual quality improvements over previous state-of-the-art diffusion-based methods on NVS tasks. 

}


\maketitle

%\saythanks % added 28-02-2014 Markus Englich

\begin{figure*}[ht!]
\begin{center}
		\includegraphics[width=2.0\columnwidth]{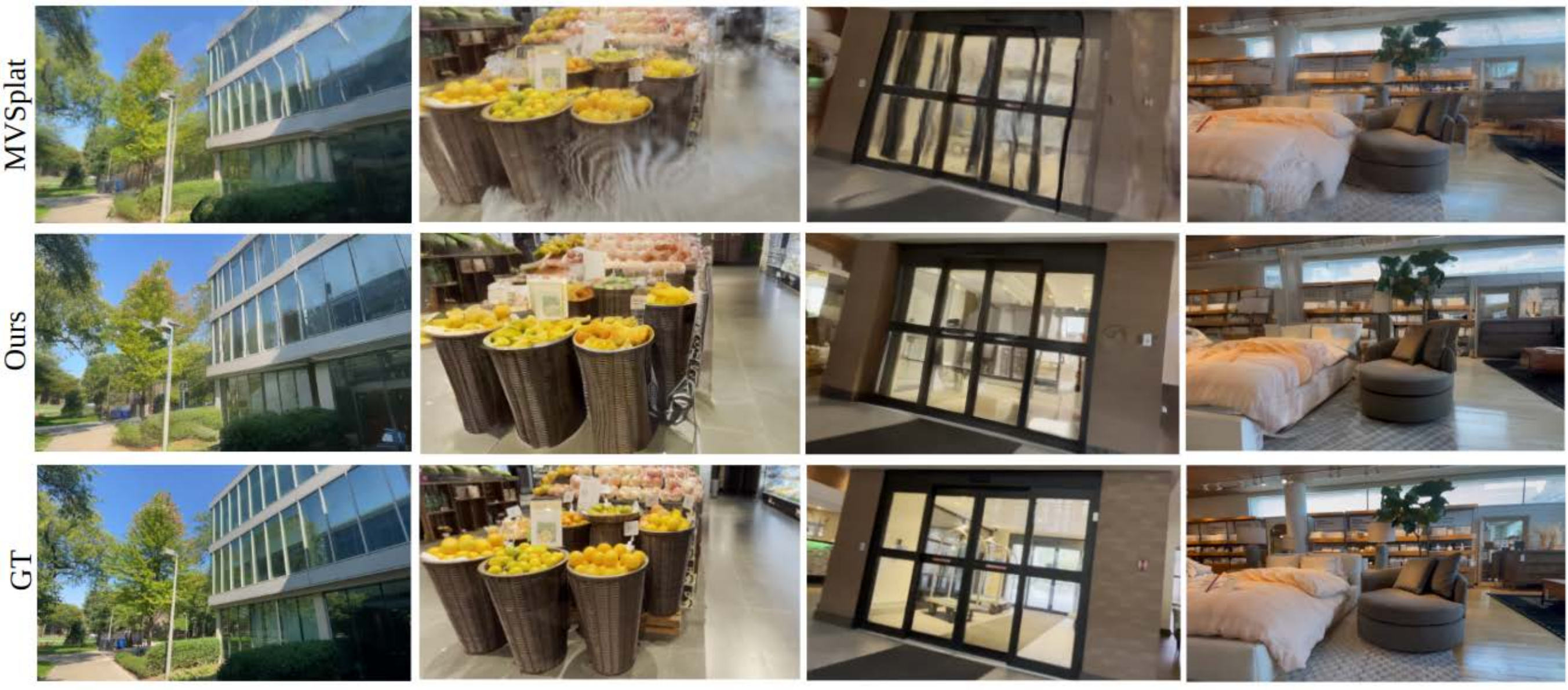}
	\caption{We demonstrate our \textbf{\method} approach on diverse in-the-wild scenes. Given sparse inputs, recent novel view synthesis methods suffer from performance degradation due to insufficient visual information. \method enhances novel view rendering quality by mitigating artifacts and recovering view-consistent details at inference time with an alias-free, representation-aligned video diffusion model.}
\label{fig:abs}
\end{center}
\end{figure*}

\section{Introduction}

Novel View Synthesis (NVS) plays a critical role in recovering 3D scenes. With the advent of Neural Radiance Fields (NeRF) \cite{mildenhall2020nerf} and 3D Gaussian Splatting (3DGS) \cite{kerbl20233dgs}, we can now render photorealistic views of complex scenes efficiently. Yet, both NeRF and 3DGS suffer from performance degradation in sparse-view settings, particularly in under-observed areas for scene-level view synthesis, which hampers their practical applicability in real-world scenarios.

To tackle this ill-posed challenge, many methods incorporate additional regularizations during the training of NeRF or 3DGS, such as cost volumes \cite{Chen2024MVSplatE3}, depth priors \cite{xu2024depthsplat} \cite{Deng2021DepthsupervisedNF} \cite{Li2024DNGaussianOS}\cite{Wang2023SparseNeRFDD} \cite{Roessle2021DenseDP} or visibility \cite{Somraj2023ViPNeRFVP} \cite{Kwak2023GeCoNeRFFN}. Despite their improvements in rendering quality of NVS, these methods still exhibit significant artifacts including spurious geometry and missing regions. Fortunately, recent advancements in video generative models pretrained on internet-scale datasets demonstrate promising capabilities in generating sequences with plausible 3D structure \cite{blattmann2023stablevideodiffusionscaling}\cite{blattmann2023videoldm}. Researchers \cite{Liu20243DGSEnhancerEU} \cite{Luo20243DEnhancerCM}\cite{wu2025difix3d}\cite{wang2025videoscenedistillingvideodiffusion}\cite{Wu2023ReconFusion3R}\cite{Chen2024MVSplat360F3} have employed diffusion models as effective enhancers for NVS from sparse views, capable of ``imagining'' unobserved regions and mitigating artifacts. Despite these improvements, these methods have limitations, particularly in two aspects: (1) lack of shift stability, and (2) limited ability to hallucinate plausible detailed appearance in underconstrained regions. Meanwhile, it is worth noting that most contemporary diffusion-based NVS enhancement methods primarily focus on optimizing solely the denoising module, specifically the U-Net denoiser architecture in video diffusion pipelines. However, the potential of diffusion models' latent representations for NVS enhancement remains unexplored.

\begin{figure*}[tb!]
\begin{center}
    \includegraphics[width=2.0\columnwidth]{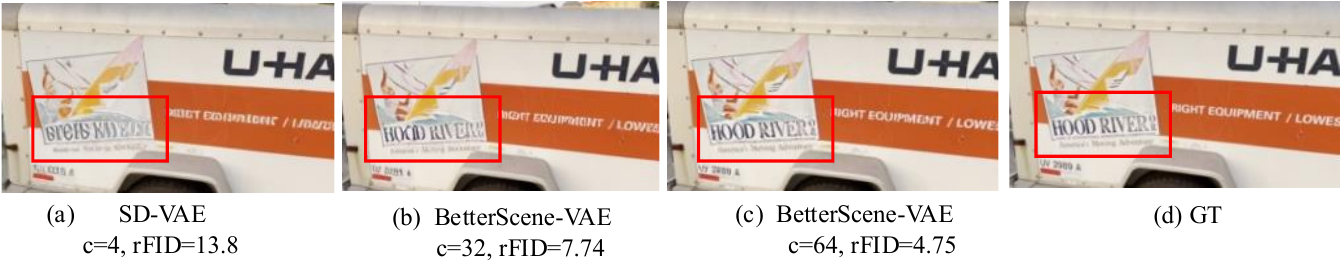}
    \caption{The visual quality and reconstruction FID score (rFID) for autoencoders with different channel sizes. We trained all the autoencoders on the DL3DV-10K \cite{ling2024dl3dv} dataset. Results show that the original 4-channel autoencoder design  \cite{Rombach2021HighResolutionIS}, which is widely used in diffusion models is unable to reconstruct fine details. Moreover, as shown in (b) and (c), increasing channel size leads to much better reconstructions. We choose to use a 64-channel \method autoencoder for our video diffusion model.}
    \label{fig:fig2}
\end{center}
\end{figure*}

In this work, we exploit the capabilities of unconstrained high-dimensional latent space for enhancing 3D scene synthesis. Several influential works \cite{blattmann2023stablevideodiffusionscaling} \cite{Dai2023EmuEI} have demonstrated that under the same spatial compression rate (or "down-sampling rate"), increasing the dimension of latent visual tokens leads to better reconstruction quality (see~\cref{fig:fig2}). This plays a key role in maintaining scene realism when using generative models as enhancers for NVS, avoiding over-hallucination while achieving higher-quality detail reconstruction. However, research \cite{blattmann2023stablevideodiffusionscaling} \cite{Xie2024SANAEH} also revealed an optimization dilemma: while increasing token feature dimensions improves reconstruction, it significantly degrades generation performance. Common strategies to address this issue include either scaling up model parameters as demonstrated by Stable Diffusion 3 \cite{blattmann2023stablevideodiffusionscaling} or sacrificing reconstruction quality with limited token dimensions. However, neither approach is suitable for NVS tasks. We argue that both the reconstruction and generative capability of latent diffusion models (LDMs) are crucial for tackling the aforementioned limitations of conventional NVS methods. Moreover, the video diffusion backbone inherently constrains model scaling. 

In this paper, building on representation-aligned LDM \cite{yu2025repa} \cite{yao2025vavae}, we propose \method, a novel view synthesis framework that incorporates feed-forward Gaussian Splatting with a representation-aligned and equivariance-regularized video diffusion model \cite{Kouzelis2025EQVAEER} \cite{zhou2025afldm}. Our key idea is to leverage high-dimensional equivariant latent representations for video LDM, achieving both superior reconstruction and generation quality to enable enhanced novel view synthesis while addressing the aforementioned limitations. Specifically, we first train a variational autoencoder (VAE) guided by vision foundation models using both an alignment loss \cite{yao2025vavae} and an equivariance loss that penalizes discrepancies between reconstructions of transformed latent representations and the corresponding input image transformations \cite{Kouzelis2025EQVAEER}. We choose Stable Video Diffusion (SVD) \cite{blattmann2023videoldm} as the enhancer backbone, integrating our pretrained VAE module and fine-tuning the denoising UNet in the second stage. Furthermore, we leverage the feed-forward 3DGS model, MVSplat \cite{Chen2024MVSplatE3}, to generate coarse novel views as SVD conditioning frames, bypassing the computationally expensive per-scene optimization required by conventional 3DGS approaches.

We evaluate our \method on the real-world scene-level DL3DV-10K dataset. Extensive results demonstrate that \method surpasses existing LDM-based NVS baselines in both fidelity and visual quality, yielding more photorealistic rendering outputs. Our main contributions can be summarized as follows.
\begin{itemize}
  \item We propose an effective framework that combines feed-forward 3D Gaussian Splatting with a representation-aligned, equivariance-regularized video LDM for novel view synthesis.  
  \item We exploit the capabilities of unconstrained high-dimensional latent spaces by training a variational autoencoder under the guidance of vision foundation models with both alignment and equivariance losses. By integrating our VAE with the SVD refinement module, we achieve enhanced reconstruction and generation quality while addressing limitations of traditional NVS methods.
  \item We conduct extensive experiments on the large-scale DL3DV-10K dataset, which contains unbounded real scenes. Results demonstrate our method's superiority over existing state-of-the-art diffusion-based NVS approaches.
\end{itemize}

\section{Related Work}
\noindent\textbf{Radiance fields novel view synthesis.} Two standard techniques that revolutionized the field of novel view synthesis are NeRF \cite{mildenhall2020nerf} and 3D Gaussian Splatting \cite{kerbl20233dgs}. NeRF utilizes an MLP to implicitly model the scene as a function and leverages volume rendering to generate novel views. Despite its high rendering quality, NeRF suffers from long training and inference times compared to 3DGS. In contrast to NeRF, 3DGS explicitly represents scenes as a set of Gaussian primitives, which are rendered to screen space through splatting-based rasterization. 3DGS offers significantly higher efficiency and competitive rendering quality compared to NeRF. However, all of these methods require high-quality, dense input views to optimize the model representation, introducing limitations in many situations. To address this, various regularization terms have been introduced to per-scene optimization ~\cite{Niemeyer2021RegNeRFRN, Li2024DNGaussianOS,Yu2022MonoSDFEM}, while others focus on speeding up the optimization process or proposing effective scene representations ~\cite{Chen2022TensoRFTR, Yu2021PlenoxelsRF, Yu2021PlenOctreesFR}. However, despite these improvements, these methods still lack generalization ability to unseen data.

\noindent\textbf{Generalizable novel view synthesis.} To avoid expensive per-scene optimization, feed-forward methods have been proposed to generate 3D representations directly from only a few input images ~\cite{Chen2024MVSplatE3, Charatan2023PixelSplat3G, Wewer2024latentSplatAV}. PixelSplat \cite{Charatan2023PixelSplat3G} predicts a dense probability distribution over 3D and generates Gaussian features from that probability distribution for scene rendering. LatentSplat \cite{Wewer2024latentSplatAV} predicts semantic 3D Gaussians in latent space, which are decoded through a lightweight generative 2D architecture. MVSplat \cite{Chen2024MVSplatE3} introduces a cost volume as a geometric constraint to enhance multi-view feature extraction, while effectively capturing cross-view feature correlations for robust depth estimation. Splatt3r \cite{Smart2024Splatt3RZG}   utilizes the foundation 3D geometry reconstruction method, MASt3R, to predict 3D Gaussian Splats without requiring any camera parameters or depth information.
While these models generate photorealistic results for observed viewpoints, their ability to reconstruct high-fidelity details in occluded or unobserved regions remains limited.

\noindent\textbf{Novel view synthesis with diffusion priors.} Recently, leveraging diffusion priors for aiding or enhancing novel view synthesis has proven to be an effective approach to improving rendering quality. By mitigating artifacts and hallucinating missing details, these methods significantly enhance the quality of synthesized views ~\cite{Chen2024MVSplat360F3, wang2025videoscenedistillingvideodiffusion, Liu2024ReconXRA, Liu20243DGSEnhancerEU, Wu2023ReconFusion3R}. ReconFusion \cite{Wu2023ReconFusion3R} fine-tunes a diffusion model on a mixture of real-world and synthetic multi-view image datasets and employs it to regularize a standard NeRF reconstruction process in a manner akin to Score Distillation Sampling. VideoScene \cite{wang2025videoscenedistillingvideodiffusion}  introduces a 3D-aware leapflow distillation strategy to bypass low-information diffusion steps. Their method enables single-step 3D scene generation. DIFIX3D+ \cite{wu2025difix3d} also allows one-step scene generation with the benefit of a consistent generative model. Furthermore, it progressively refines the 3D representation by distilling back the enhanced views to achieve significant results. 3DGSEnhancer \cite{Liu20243DGSEnhancerEU} employs video diffusion to restore view-consistent novel view renderings, then utilizes these refined views to optimize the initial 3DGS model. MVSplat360\cite{Chen2024MVSplat360F3} leverages a feed-forward 3DGS model to directly generate coarse geometric features in the latent space of a pre-trained SVD model, enabling efficient synthesis of photorealistic, wide-sweeping novel views. While our approach builds upon MVSplat360's pipeline, we introduce an innovative representation-aligned, equivariance-regularized high-dimensional latent feature representation instead of using an off-the-shelf pretrained SVD. Our experiments demonstrate superior fidelity and visual quality compared to baseline methods.

\section{Methodology}
\label{sec:method}

\subsection{\method Overview} 

\begin{figure*}[tb!]
    \centering
    \includegraphics[width=\textwidth]{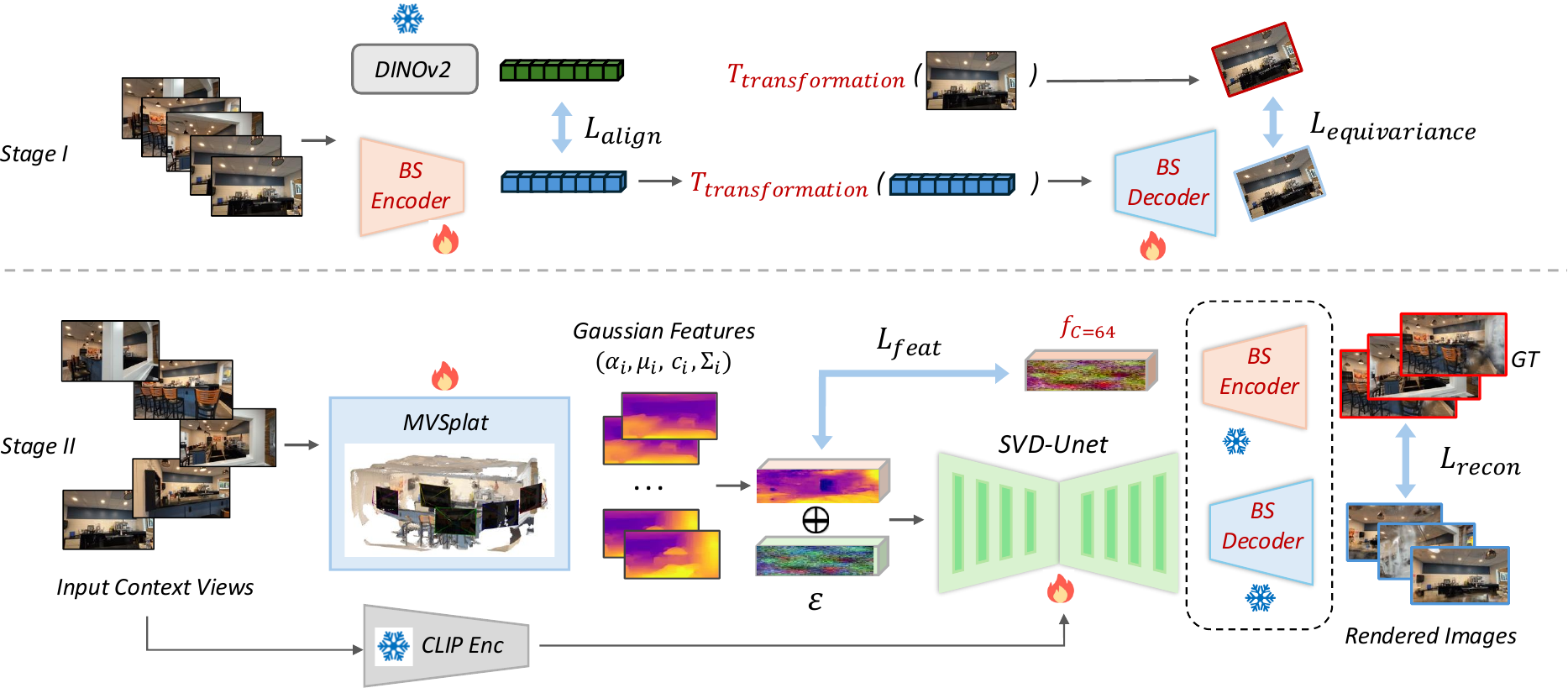}
    \caption{\textbf{Overview of our \method}. The training process consists of two stages. In the first stage, we train an autoencoder using a representation-aligned and equivariance-regularized objective function. In the second stage, we freeze the pretrained \method-VAE and fine-tune the denoiser U-Net within the SVD framework. We leverage a feed-forward 3DGS rendering module, MVSplat, to generate both coarse synthesized views and corresponding Gaussian feature latents. The SVD module then processes these coarse features to decode enhanced high-quality images. }
    \label{fig:framework}
\end{figure*}

\method consists of a feed-forward 3DGS reconstruction module, MVSplat \cite{Chen2024MVSplatE3}, and a refinement module based on a stable video diffusion \cite{blattmann2023videoldm} backbone. Specifically,  given $N$ sparse-view inputs $\mathcal{I}=\{\bm{I}^i\}_{i=1}^N$, our goal is to synthesize realistic images from novel viewpoints in an end-to-end manner. The framework of our \method is illustrated in ~\cref{fig:framework}. The training of our \method consists of two stages. In the first stage, we train an autoencoder using a representation-aligned and equivariance-regularized objective function. In the second stage, we freeze the pretrained \method-VAE and fine-tune the denoiser U-Net within the SVD framework. As shown in \cref{fig:framework}, we leverage a feed-forward 3DGS rendering module, MVSplat, to generate both coarse synthesized views and corresponding Gaussian feature latents $\hat{\bm{f}}_i$. The SVD module then processes these coarse features to decode enhanced high-quality images. Further details are discussed in subsequent sections.

\subsection{Representation-aligned Equivariance-regularized VAE} 

In this section, we introduce the representation-aligned and equivariance-regularized variational autoencoder for achieving superior quality in both reconstruction and generation. This optimization improves both synthesis fidelity and visual quality of NVS by incorporating unconstrained high-dimensional latent representations into the SVD pipeline. Specifically, we scale the original SD-VAE architecture which uses 8× spatial downsampling and 4 latent channels, to 16× downsampling with 64 latent channels that maintain a comparable model scale. This modification triggers the aforementioned optimization dilemma: while reconstruction quality improves, generation performance degrades. This phenomenon likely stems from the Gaussian prior assumption in the VAE's KL divergence loss. The objective function of the original VAE \cite{kingma2022autoencodingvariationalbayes} derived from maximizing the evidence lower bound (ELBO), can be expressed as:
\begin{align*}
\LB{}{\bxi} = - D_{KL}(\qPhi(\bz|\bxi) || \pT(\bz)) \\
    + \Exp{\qPhi(\bz|\bxi)}{\log \pT(\bxi | \bz)}
\label{eq:lowerbound2}\eqnr\end{align*}
where the posterior $\qPhi(\bz|\bxi)$ is constrained to match a standard Gaussian distribution. This inherently restricts latent embedding expressiveness, especially in high-dimensional spaces. Another observation is that increasing the latent dimension leads to underutilization of the feature space, which is also observed in autoregressive generation with codebook embeddings. ~\cite{Bo2024EnhancingCU, Zhu2024ScalingTC}. 

\noindent\textbf{Representation alignment loss.} To generate a high-dimensional latent space for enhanced novel view synthesis (NVS), we introduce a vision foundation model alignment loss ~\cite{yu2025repa, yao2025vavae} to optimize the VAE component within the original SVD framework. The key idea involves constraining the latent space by leveraging the vision foundation model's feature space. This enables a flexible feature distribution that improves feature utilization while escaping the limitations of the standard Gaussian distribution assumption. 

Specifically, given an input image \( I \), we process it through both our modified VAE with 64 latent channels and DINOv2 \cite{Oquab2023DINOv2LR}, a vision foundation model that extracts robust visual features. The resulting image latents are denoted as \( Z_{V} \) and \( F_{D} \). \( Z_{V} \) is projected to match the dimensionality of \( F_{D} \) through a linear transformation: \( Z' \) = W \( Z_{V} \). We employ the cosine similarity loss \cite{yao2025vavae} to minimize the discrepancy between corresponding feature representations with margin \( m_1 \).
\begin{equation}
\label{eq:margin_cos}
\mathcal{L}_{\text{cos-align}} = \frac{1}{h \times w} \sum_{i=1}^{h} \sum_{j=1}^{w} \text{ReLU} \left( 1 - m_1 - \frac{z'_{ij} \cdot f_{ij}}{\|z'_{ij}\| \|f_{ij}\|} \right)
\end{equation}
Furthermore, a distance similarity loss is employed as a complementary objective to align the internal distributions of \( Z_{V} \) and \( F_{D} \) with margin \( m_2 \).
\begin{equation}
\label{eq:dist_matrix_sim_margin}
\mathcal{L}_{\text{dist-align}} = \frac{1}{N^2} \sum_{i,j} \text{ReLU} \left(\left| \frac{z_i \cdot z_j}{\|z_i\| \|z_j\|} - \frac{f_i \cdot f_j}{\|f_i\| \|f_j\|} \right| - m_2 \right)
\end{equation}

\noindent\textbf{Equivariance regularization.} Recent research reveals that the SD-VAE latent representations lack equivariance under spatial transformations. Specifically, given an input image \( I \) and its corresponding VAE latent \( Z_{V}( I ) \), if we apply a transformation $\tau$ to both \( I \) and \( Z_{V}( I ) \), the latent representation of the transformed image should satisfy \cite{Kouzelis2025EQVAEER}: 
\begin{align}
    \label{eq:equivariace}
    \forall \mathbf{I} \in \mathcal{I}: \quad \mathcal{Z}(\tau \circ \mathbf{I}) = \tau \circ \mathcal{Z}(\mathbf{I})\text{.}
\end{align}
Violating this property implicitly leads to temporal inconsistency in video LDMs, as the noise patterns between frames lack transformation consistency. Consequently, the decoded images cannot form an equivariant frame sequence, resulting in sudden scene shifts or inconsistent content across consecutive frames. This creates fundamental limitations for using video LDMs to enhance novel view synthesis (NVS), which requires strict temporal consistency.

Therefore, we directly enforce latent equivariance by incorporating the constraint from \eqref{eq:equivariace} as a regularization term during autoencoder training with augmented transformations.
\begin{equation}
    \label{eq:encoder_loss}
     \mathcal{L}_{\text{latent-equivariance}}(\mathbf{I}) = 
     % \underset{\substack{\tau \sim \mathcal{T}}}{\mathbb{E}}\big[ 
     \Vert \tau \circ \mathcal{Z}(\mathbf{I}) - \mathcal{Z}(\tau \circ  \mathbf{I}) \Vert_2^2 
     % \big]
     \text{,}
\end{equation}
where $\tau$ represents a set of spatial transformations. 
In addition to the latent equivariance loss, we employ a reconstruction equivariance loss to align the reconstructions of transformed latent features $(\mathcal{D}( \tau \circ \mathcal{Z}(\mathbf{I}) ))$ with the corresponding transformed inputs ($\tau \circ \mathbf{I}$ ). The reconstruction equivariance objective is as follows:
\begin{align}
    \label{eq:ours_obj}
    \mathcal{L}_{\text{recon-equivariance}} (\mathbf{I}, {{\tau}}) = 
     \mathcal{L}_{rec}& \Big({{\text{$\mathbf{\tau} \circ$}}} \mathbf{I}, \mathcal{D}\big ({{\text{$\mathbf{\tau} \circ$}}}\mathcal{Z}(\mathbf{I}) \big) \Big) + \\
    \lambda_{gan} \mathcal{L}_{gan}& \Big( \mathcal{D}\big ({{\text{$\mathbf{\tau} \circ$}}} \mathcal{Z}(\mathbf{I}) \big) \Big) + \lambda_{reg} \mathcal{L}_{reg} \notag
    % \Big] 
\end{align}

\noindent\textbf{\method-VAE.} We train our autoencoder on the DL3DV-10K dataset with the objective function:
\begin{equation}
\begin{split}
\mathcal{L}_{\text{\method-VAE}} = w_{\text{align}} *(\mathcal{L}_{\text{dist-align}} + \mathcal{L}_{\text{cos-align}}) + \\
w_{\text{equi}} * (\mathcal{L}_{\text{latent-equivariance}} +  \mathcal{L}_{\text{recon-equivariance}}) 
\end{split}
\end{equation}
By leveraging representation alignment and equivariance regularization, our autoencoder achieves both superior reconstruction fidelity and generation capability while enabling transformation equivariance in the latent space. By integrating this high-dimensional latent representation into SVD refinement modules, we produce enhanced visual fidelity and rendering quality for NVS.

\subsection{\method: Video LDM NVS Enhancer}
\label{sec:betterscene}

Given coarse rendered novel views with artifacts $\tilde{\mathcal{I}}$, our model generates a sequence of cleaned predictions. We build our pipeline upon a feed-forward 3DGS reconstruction model, MVSplat, and a pretrained SVD backbone. We introduce the details of each module in the following parts.

\noindent\textbf{Coarse feature generation.} 
We bypass expensive per-scene optimization ~\cite{Barron2021MipNeRFAM,Barron2021MipNeRF3U,Gao2024CAT3DCA}, and adopt MVSplat, a generalizable feed-forward 3DGS generation model capable of synthesizing novel views for unseen scenes from sparse-view inputs. Specifically, MVSplat first fuses multi-view information and obtains cross-view aware features $\mathcal{F}=\{\bm{F}^i\}_{i=1}^N$ given sparse-view observations $\mathcal{I}=\{\bm{I}^i\}^{N}_{i=1}$ and their corresponding camera poses $\mathcal{P}=\{\bm{P}^i\}_{i=1}^N$.
Then, $N$ cost volumes $\mathcal{C}=\{\bm{C}^i\}_{i=1}^N$ %
are constructed through cross-view feature correlation matching, enabling per-view depth estimation. Finally, we compute the Gaussian parameters: mean $\bm{\mu}$, covariance $\Sigma\in\mathbb{R}^{3\times3}$, and spherical harmonic coefficients $\bm{c}\in\mathbb{R}^{3(S+1)^2}$ where $S$
is the order. The target view $\tilde{\mathcal{I}}$ can be rendered through rasterization. 

\noindent\textbf{Gaussian feature conditioning.} In the original SVD framework, the first ground truth frame usually serves as the conditioning input, which is concatenated with Gaussian noise during the denoising process. In our framework, we leverage coarse rendered priors by directly concatenating rasterized features $\hat{\mathcal{F}}$ as conditioning with the latent space noise in SVD. Notably, the coarse Gaussian priors are directly combined with noise latents, bypassing the encoding step, similar to the method described in \cite{Chen2024MVSplat360F3}. The key advantage of this operation is its ability to leverage supervision from ground truth frame VAE embeddings, which simultaneously optimizes both the conditioning Gaussian features and the MVSplat modules. This is where our optimized VAE plays a critical role in the pipeline. The high-dimensional expressive latent representations of target frames provide strong supervision for the latent conditioning, thereby offering more effective guidance for the generation process. 

Moreover, similar to the original SVD, we leverage CLIP~\cite{Radford2021LearningTV} to generate conditioning embeddings from the input views $\mathcal{I}$. These embeddings serve as global semantic cues that are injected into the denoising process via cross-attention operations, helping the model maintain both semantic coherence and fidelity.

\noindent\textbf{Fine-tuning and losses.} We fine-tune the denoiser U-Net of SVD while keeping our pretrained VAE encoder and decoder frozen. The model takes sparse context images as input and generates refined target images as output through our \method framework. The entire system is trained end-to-end. We supervise our SVD model with: (1) the standard v-prediction formulation as the diffusion loss, and (2) a linear combination of $\ell_2$ and LPIPS~\cite{Zhang2018TheUE} discrepancies between the predicted outputs $\tilde{\mathcal{I}}^\mathrm{pred}$ and the corresponding ground truth $\mathcal{I}^\mathrm{gt}$ as the reconstruction loss. Additionally, as previously mentioned, we align the conditioning Gaussian features with the high-dimensional latent representations of target images encoded by our pretrained VAE with a latent feature loss
$
    \min_{g_\theta}
    \mathbb{E}_{\hat{\mathcal{Z}}\sim g(\mathcal{I})}
    \lVert
    \mathcal{E}(\mathcal{I}^\mathrm{gt}) - \hat{\mathcal{Z}}^\mathrm{gs}
    \rVert^2_2.
$

\section{Experiments}

\subsection{Implementation Details}

To validate the efficacy of \method, we conduct experiments on the challenging DL3DV-10K dataset, which contains 51.3 million frames from 10,510 real-world scenes. Our experiments follow the same benchmark settings as \cite{Chen2024MVSplat360F3}. The test partition contains 140 scenes and is filtered from the training set. We select 5 input views and evaluate 56 novel views, sampled uniformly from the remaining frames. We fine-tune the original SVD that generates 14 frames per sampling epoch. Our autoencoder architecture employs a 16× downsampling rate and a latent channel size of 64. The entire pipeline is trained on four NVIDIA H100 GPUs.

\begin{table}[tb!] \caption{\textbf{A quantitative comparison of novel view synthesis performance using 5 input views}.
    Experiments on the DL3DV-10K dataset follow the setting of \cite{Chen2024MVSplat360F3} (Note that since all experimental settings remain identical, we directly adopt the evaluation results for baseline methods reported in \cite{Chen2024MVSplat360F3}.)}
    \label{tab:sota_dl3dv}
    \centering
    \setlength\tabcolsep{2pt}
    \resizebox{0.48\textwidth}{!}{\begin{tabular}{@{}l ccccc c ccccc@{}}
         \toprule
         \multirow{1}{*}{\textbf{Method}} 
         &  PSNR$\uparrow$ & SSIM$\uparrow$ &  LPIPS$\downarrow$ & FID$\downarrow$   \\
         \midrule
         MVSplat~\cite{Chen2024MVSplatE3}            & 17.05 & 0.499 &  0.435  & 61.92 \\
         latentSplat~\cite{Wewer2024latentSplatAV}   & 17.79 & 0.527 &  0.391  & 34.55 \\
         MVSplat360~\cite{Chen2024MVSplat360F3}   & \textbf{17.81} & 0.562 & 0.352  & 18.89
         \\
         \method (ours)        & \textbf{17.81} & \textbf{0.579} & \textbf{0.347}  & \textbf{16.59} &&
         \\
         \bottomrule
    \end{tabular}
   }
\end{table}

\begin{figure*}[tb!]
    \centering
    \includegraphics[width=1.0\textwidth]{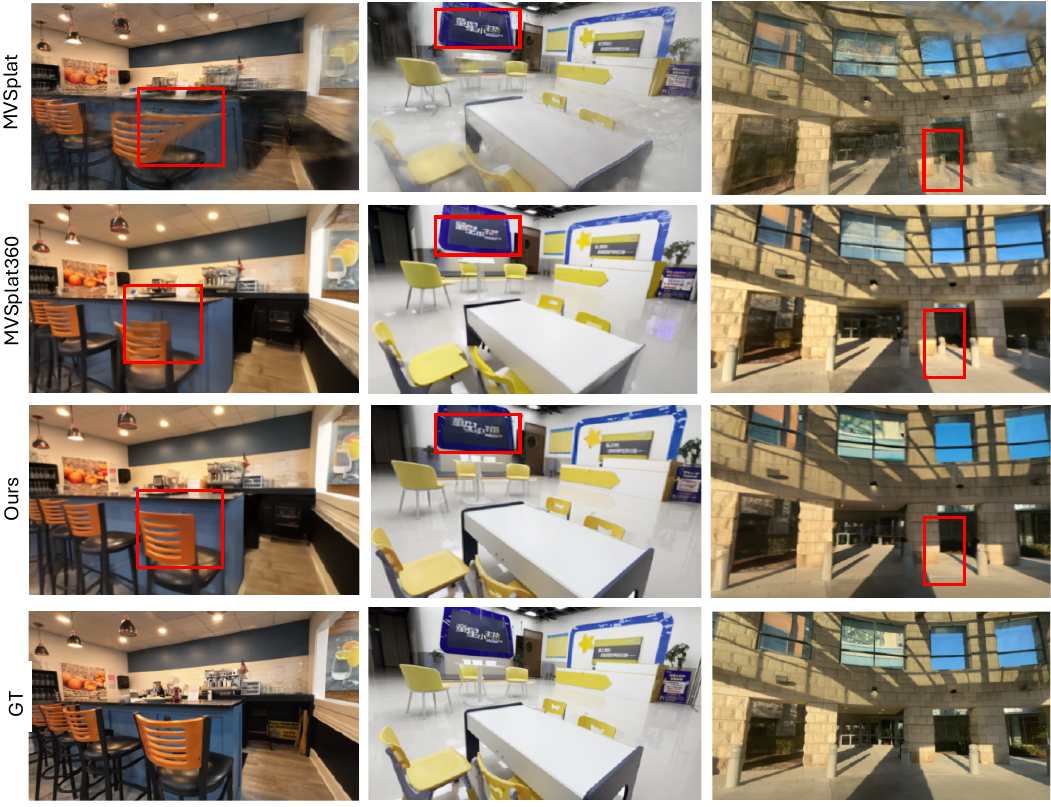}
    \begin{picture}(0,0)
    \end{picture}
    \vspace{-1.5em}
    \caption{A visual comparison of enhanced rendering results generated from 5 input views across scenes from the DL3DV benchmark test set. \method demonstrates superior visual quality and enhanced detail consistency compared to existing state-of-the-art approaches.}
    \vspace{-0.5em}
    \label{fig:comparisons}
\end{figure*}

% \begin{figure*}[tb!]
%     \centering
%     \includegraphics[width=2\columnwidth]{figures/fig_ablation.pdf}
%     \begin{picture}(0,0)
%     \end{picture}
%     \caption{ A qualitative comparison of reconstruction performance across channel sizes. The SD-VAE represents the original VAE architecture with 4 latent channels.}
%     \label{fig:ablation}
% \end{figure*}

\subsection{Comparison with State-of-the-Arts}

The quantitative results with 5 input views on the DL3DV test set are shown in \cref{tab:sota_dl3dv}. Our approach achieves superior performance compared to all baseline methods in SSIM, LPIPS, and FID metrics, while maintaining PSNR scores comparable to MVSplat360. 

The qualitative results on the DL3DV-10K benchmark are presented in \cref{fig:comparisons}.  We compare our \method with MVSplat \cite{Chen2024MVSplatE3} and its diffusion-enhanced variant, MVSplat360 \cite{Chen2024MVSplat360F3}. Without the diffusion-based refinement, MVSplat generates blurry novel views due to insufficient constraints from sparse input views. With the refinement from video diffusion, MVSplat360 demonstrates significant improvement, achieving remarkable visual quality through effective artifact removal. However, imperfections persist in both the reconstructed geometry and detail consistency. The first column in \cref{fig:comparisons} demonstrates \method's capability for effectively removing artifacts. The second and third columns in \cref{fig:comparisons} validate: (1) the efficacy of high-dimensional latent representations for improved recovery in images, such as text on the wall, and (2) the effectiveness of our representation-aligned, equivariance-regularized autoencoder design for maintaining detail consistency.
Overall, our approach outperforms all baseline methods in both visual quality and detail consistency, demonstrating the capability to synthesize high-fidelity novel views.

\subsection{Ablations Study}

The core innovation of our pipeline is the high-dimensional latent feature representation. In this section, we present an ablation study exploring the impact of latent channel size on our \method-VAE performance. Due to the prohibitive computational cost of training the complete \method pipeline with SVD on the full DL3DV-10K dataset, we focus our evaluation on the reconstruction performance of \method-VAE across three latent channel configurations: 16, 32, and 64 dimensions. 

As demonstrated in \cref{tab:ablations}, increasing the latent dimensionality yields significant improvements in reconstruction quality. The 64-channel configuration achieves superior performance across all metrics compared to lower-dimensional latent representations. Notably, higher-dimensional representations consistently produce superior reconstructions, with the 64-channel configuration achieving particularly robust detail consistency. These results suggest potential reasons for \method's superior performance in high-frequency detail and complex texture enhancement compared to existing approaches.

\begin{table}[tb!] \caption{A quantitative comparison of reconstruction performance across latent channel sizes. The SD-VAE represents the original VAE architecture with 4 latent channels.}
    \label{tab:ablations}
    \centering
    \setlength\tabcolsep{2pt}
    \resizebox{0.48\textwidth}{!}{\begin{tabular}{@{}l ccccc c ccccc@{}}
         \toprule
         \multirow{1}{*}{\textbf{Method}} 
         &  PSNR$\uparrow$ & SSIM$\uparrow$ &  LPIPS$\downarrow$ & rFID$\downarrow$   \\
         \midrule
         SD-VAE           & 26.06 & 0.78 & 0.12  & 13.83 \\
         \method-VAE(C=16)  & 25.14 & 0.758 &  0.12  & 13.41 \\
         \method-VAE(C=32)   & 28.21 & 0.851 & 0.08  & 7.41
         \\
         \method-VAE(C=64)     & \textbf{31.21} & \textbf{0.92} & \textbf{0.04}  & \textbf{4.90} &&
         \\
         \bottomrule
    \end{tabular}
   }
\end{table}

\section{Conclusion}
\label{sec:conclusion}

We present \method, an approach for enhancing novel view synthesis (NVS) quality from sparse and unconstrained photo collections. Unlike contemporary methods, we investigate the diffusion model's latent space and introduce (1) equivariance regularization and (2) vision foundation model-aligned representations, both applied to the variational autoencoder (VAE) within the SVD pipeline. Our framework enhances NVS quality and generates artifact-free, temporally consistent novel views. We evaluated \method on the challenging DL3DV-10K benchmark. Our method demonstrates significant visual quality improvements over baseline approaches. Our work may contribute insights for advancing 3D reconstruction and view generation in future research. However, the SVD model in the \method framework requires computationally expensive training. Future work could explore replacing this pipeline with more efficient video diffusion architectures.

\section*{Acknowledgments}

This work was supported in part by the U.S. Army Research Office under Grant AWD-110906.

\bibliography{ISPRSguidelines_authors}
\end{document}